\documentclass[11pt]{article}
\usepackage{acl2010}
\usepackage{times}
\usepackage{latexsym}

\usepackage[T1]{fontenc}
\usepackage[utf8]{inputenc}
\usepackage{graphicx}
\usepackage{tabu}
\usepackage{xcolor}
\usepackage{ragged2e}
\usepackage{placeins}
\usepackage{tabu}
\usepackage{url}
\usepackage{latexsym}
\usepackage{booktabs}
\usepackage{amsmath,amsfonts}
\usepackage{braket}
\usepackage{multirow}
\usepackage{multicol}
\usepackage{titlesec}
\usepackage{relsize}
\usepackage[breaklinks=true]{hyperref}
\usepackage{colortbl}
\usepackage{lipsum}
\usepackage{fancyhdr}
\usepackage{enumitem}
\usepackage[flushmargin]{footmisc}
\usepackage{setspace}

\usepackage{caption}

\newcommand{\tit}[1]{\textit{#1}}
\newcommand{\tbf}[1]{\textbf{#1}}

\newcommand{\citep}[1]{\cite{#1}}
\newcommand{\citet}[1]{\newcite{#1}}

\title{Sense Vocabulary Compression through the Semantic Knowledge of WordNet for Neural Word Sense Disambiguation}

\author{
    \vspace{5.0pt}
    Loïc Vial \quad Benjamin Lecouteux \quad Didier Schwab\\
    \vspace{5.0pt}
        Univ. Grenoble Alpes, CNRS, Grenoble INP, LIG, 38000 Grenoble, France\\
        \texttt{\{loic.vial, benjamin.lecouteux, didier.schwab\}@univ-grenoble-alpes.fr}
}

\date{}

\begin{document}

\maketitle

\begin{abstract}
In this article, we tackle the issue of the limited quantity of manually sense annotated corpora for the task of word sense disambiguation, 
by exploiting 
the semantic relationships between senses such as synonymy, hypernymy and hyponymy, in order to compress the sense vocabulary of Princeton WordNet, and thus reduce the number of different sense tags that must be observed to disambiguate all words of the lexical database. 
We propose two different methods that 
greatly reduce 
the size of neural WSD models, 
with the benefit of improving their coverage without additional training data, and without impacting their precision. 
In addition to our methods, we present 
a WSD system which relies on pre-trained BERT word vectors 
in order to achieve results 
that significantly outperforms the state of the art on all WSD evaluation tasks. 
\end{abstract}

\section{Introduction}

Word Sense Disambiguation (WSD) is a task which aims to clarify a text by assigning to each of its words the most suitable sense labels, given a predefined sense inventory.

Various approaches have been proposed to achieve WSD:
Knowledge-based methods rely on dictionaries, lexical databases, thesauri or knowledge graphs as primary resources, and use algorithms such as lexical similarity measures \citep{Lesk1986} or graph-based measures \citep{Moro2014EntityLM}. Supervised methods, on the other hand, exploit sense annotated corpora as training instances for a 
classifier such as SVM \citep{Chan2007,Zhong2010}, or more recently by a neural network \citep{kaageback2016word}. Finally, unsupervised methods automatically identify the different senses of words from unannotated or parallel corpora (e.g. \citet{ide2002}). 

Supervised methods are by far the most predominant as they generally offer the best results in evaluation campaigns (for instance \citep{Navigli2007}). State of the art classifiers used to combine 
specific features such as the parts of speech and the lemmas of surrounding words
\citep{Zhong2010}, 
but they are now replaced by 
neural networks which learn their own representation of words \citep{raganato2017,minh2018}.

One major bottleneck of supervised systems is the restricted quantity of manually sense annotated corpora: In the annotated corpus SemCor \citep{Miller1993}, the largest manually sense annotated corpus available, words are annotated with 33\,760 different sense keys, which corresponds to only approximately 16\% of the sense inventory of
WordNet \citep{miller1995wordnet},
the lexical database of reference widely used in WSD.
Many works try to leverage this problem by creating new sense annotated corpora, either automatically \citep{pasini2017}, semi-automatically \citep{taghipourng2015}, or through crowdsourcing \citep{yuan_2016}. 

In this work, the idea is to solve this issue by taking advantage of the semantic relationships between senses included in WordNet, such as the hypernymy, the hyponymy, the meronymy, the antonymy, etc. Our method is based on the 
observation that a sense and its closest related senses (its hypernym or its hyponyms for instance) all share a common idea or concept, 
and so a word can sometimes be disambiguated using only related concepts. Consequently, we do not need to know every sense of WordNet to disambiguate all words of WordNet.

For instance, let us consider the word ``mouse'' and two of its senses which are the \tit{computer} mouse and the \tit{animal} mouse. We only need to know the notions of ``animal'' and ``electronic device'' to distinguish them, and all notions that are more specialized such as ``rodent'' or ``mammal'' are therefore superfluous. By grouping them, we can benefit from all other instances of electronic devices or animals in a training corpus, even if they do not mention the word ``mouse''.

\noindent\tbf{Contributions:}
In this paper, we hypothesize that only a subset of WordNet senses could be considered to disambiguate all words of the lexical database. Therefore, we propose two different methods for building this subset and we call them sense vocabulary compression methods. 
By using these techniques, we are able to greatly improve the coverage 
of supervised WSD systems, nearly eliminating the need for a backoff strategy that is currently used in most systems when dealing with a word which has never been observed in the training data.
We evaluate our method on a state of the art WSD neural network,
based on pretrained contextualized word vector representations, 
and we present results that significantly outperform the state of the art on every standard WSD evaluation task. Finally, we provide a documented tool for training and evaluating neural WSD models, as well as our best pretrained model in a 
dedicated 
GitHub repository\footnote{
\url{https://github.com/getalp/disambiguate}}.

\section{Related Work}

In WSD, several recent advances have been made in the creation of new neural architectures for supervised models and the integration of knowledge into these systems. Multiple works also exploit the idea of grouping together related senses. In this section, we give an overview of these works.

\subsection{WSD Based on a Language Model}

In this type of approach, that has been initiated by \citet{yuan_2016} and reimplemented by \citet{minh2018}, the central component is a neural language model able to predict a word with consideration for the words surrounding it, thanks to a recurrent neural network trained on a massive quantity of unannotated data.

Once the language model is trained, it is used to produce sense vectors that result from averaging the word vectors predicted by the language model at all positions of words annotated with the given sense.

At test time, the language model is used to predict a vector according to the surrounding context, and the sense closest to the predicted vector is assigned to each word.

These systems have the advantage of bypassing the problem of the lack of sense annotated data by concentrating the power of abstraction offered by recurrent neural networks on a good quality language model trained in an unsupervised manner.
However, sense annotated corpora are still indispensable to contruct the sense vectors.

\subsection{WSD Based on a Softmax Classifier}\label{sec:classif_based_wsd}

In these systems, the main neural network directly classifies and attributes a sense to each input word through a probability distribution computed by a softmax function. Sense annotations are simply seen as tags put on every word, like a POS-tagging task for instance.

We can distinguish two separate branches of these types of neural networks: 
\begin{enumerate}[leftmargin=*,topsep=0pt,itemsep=0pt,parsep=4pt]
    \item Those in which we have several distinct and token-specific neural networks (or classifiers) for every different word in the dictionary \citep{iacobacci2016embeddings,kaageback2016word}, each of them being able to manage a particular word and its particular senses. For instance, one of the classifiers is specialized in choosing between the four possible senses of the noun ``mouse''. This type of approach is particularly fitted for the lexical sample tasks, where a small and finite set of very ambiguous words have to be sense annotated in several contexts, but it can also be used in all-words word sense disambiguation tasks.
    \item Those in which we have a larger and general neural network that is able to manage all different words and assign a sense in the set of all existing sense in the dictionary used \citep{raganato2017}.
\end{enumerate}

{
\noindent The advantage of the first branch of approaches is that in order to disambiguate a word, limiting our choice to one of its possible senses is computationally much easier than searching through all the senses of all words. To put things in perspective, the average number of senses of polysemous words in WordNet is approximately 3,
whereas the total number of senses considering all words is 206\,941.}

The second approach, however, has an interesting property: all senses reside in the same vector space and hence share features in the hidden layers of the network. This allows the model to predict an identical sense for two different words (i.e. synonyms), but it also offers the possibility to predict a sense for a word not present in the dictionary (e.g. neologism, spelling mistake...).

Finally, in two recent articles, \citet{luo2018b} and \citet{luo2018a} have proposed an improvement of these type of architectures, by computing an attention between the context of a target word and the gloss of its different senses. Thus, their work is one of the first to incorporate knowledge from WordNet into a WSD neural network.

\subsection{Sense Clustering Methods}

Several works exploit the idea of grouping together mutiple WordNet sense tags in order to create a coarser sense inventory which can potentially be more useful in some NLP tasks.

In the works of \citet{Ciaramita2006}, the authors propose a supervised system that learns and predicts 
``Supersense'' tags,
which belong to the set of the broad semantic categories of senses, organizing the sense inventory of WordNet. This tagset consists, in their work, of 26 categories for nouns (such as ``food'', ``person'' or ``object''), and 15 categories for verbs (such as ``emotion'' or ``weather''). 
By predicting supersense tags instead of the usual fine-grained sense tags of WordNet, the output vocabulary of their system is shrinked to only 41 different classes, and this leads to a small and easy-to-train model able to perform partial WSD, which could be useful and sufficient for other NLP tasks where the fine-grained distinction is not necessary. 

In \citet{izquierdo2007BLC}, the authors 
propose several methods for creating ``Basic Level Concepts'' (BLC),
groups of related senses with a generally smaller size than supersenses, and which can be controlled by a threshold variable. Their methods rely on the semantic relationships between senses of WordNet,
and, in the same way as \citet{Ciaramita2006}, they evaluated their clusters on a modified WSD task, where supersenses or BLC have to be predicted instead of the original sense tags from WordNet.

The main difference between our work and these works is that our end goal is to improve fine-grained WSD systems. Even though our methods generate clusters of related senses, we guarantee that two different senses of a lemma reside in two different clusters, so at the end, even if our supervised system produces a cluster tag for a target word, we are still able to find back the true sense tag,
by simply keeping track of which sense key of its lemma belongs to the predicted group.

\section{Sense Vocabulary Compression} 

Current state of the art supervised WSD systems such as \citet{yuan_2016}, \citet{raganato2017}, \citet{luo2018b} and \citet{minh2018} are all confronted to the following issues:

\begin{enumerate}[leftmargin=*,topsep=0pt,itemsep=0pt,parsep=0pt,partopsep=0pt]
    \item Due to the small number of manually sense annotated corpora available, a target word may never be observed during the training, and therefore the system is not able to annotate it.
    \item For the same reason, a word may have been observed, but not all of its senses. In this case the system is able to annotate the word, but if the expected sense has never been observed, the output will be wrong, regardless of the architecture of the supervised system.
    \item Training a neural network 
to predict a tag which belongs to the set of all WordNet senses can become 
    extremely 
    slow and requires a lot of parameters with a large output vocabulary. And this vocabulary goes up to 206\,941 if we consider all word-senses of WordNet.
\end{enumerate}

\noindent In order to overcome all these issues, we propose a method for grouping together multiple sense tags that refer in fact to the same concept.
In consequence, 
the output vocabulary decreases, the ability of the trained system to generalize improves, as well as its coverage.

\subsection{From Senses to Synsets: A Vocabulary Compression Based on Synonymy}

In the lexical database WordNet, senses are organized in sets of synonyms called synsets. A synset is technically a group of one or more word-senses that have the same definition and consequently the same meaning. For instance, the first senses of ``eye'', ``optic'' and ``oculus'' all refer to a common synset which definition is ``the organ of sight''.

\begin{figure}[htbp]
\centering
\includegraphics[width=0.9\linewidth]{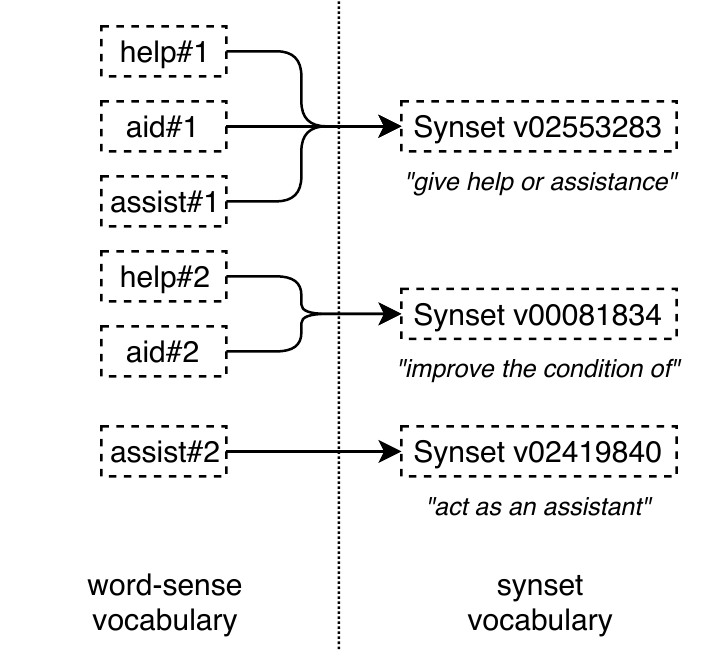}
\caption{Word-sense to synset mapping (compression through synonymy) applied on the first two senses of the words ``help'', ``aid'' and ``assist''.}
\label{fig:sense_to_synset}
\end{figure}

Illustrated in \autoref{fig:sense_to_synset}, the word-sense to synset mapping is hence a way of compressing the output vocabulary, and it is already applied in many works \citep{yuan_2016,minh2018}, while not being always explicitly stated. This method clearly helps to improve the coverage of supervised systems however. Indeed, if the verb ``help'' is observed in the annotated data in its first sense, 
the context surrounding the target word can be used to later annotate the verb ``assist'' or ``aid'' with the same valid synset tag.

Going further, other information from WordNet can help the system to generalize. 
Our first new method takes advantage of the hypernymy and hyponymy relationships to achieve the same idea.

\subsection{Compression through Hypernymy and Hyponymy Relationships}

According to \citet{polguere2003}, hypernymy and hyponymy are two semantic relationships which correspond to a particular case of sense inclusion:
the hyponym of a term is a specialization of this term, whereas its hypernym is a generalization. For instance, a ``mouse'' is a type of ``rodent'' which is in turn a type of ``animal''.

In WordNet, these relationships bind nearly every noun together in a tree structure\footnote{We computed that 41\,607 on the 44\,449 polysemous nouns of WordNet (94\%) are part of this hierarchy.} that goes from the generic root, 
the node ``entity'' to the most specific leaves, for instance 
the node ``white-footed mouse''. These relationships are also present on several verbs: for instance ``add'' is a way of ``compute'' which 
is a way of ``reason''. 

For the sake of WSD, just like grouping together the senses of the same synset helps to better generalize, we hypothesize that grouping together the synsets of the same hypernymy relationship also helps in the same way. 
The general idea of our method is that the most specialized concepts in WordNet are often superfluous for WSD. 

\begin{figure}[htbp]
\centering
\includegraphics[width=0.9\linewidth]{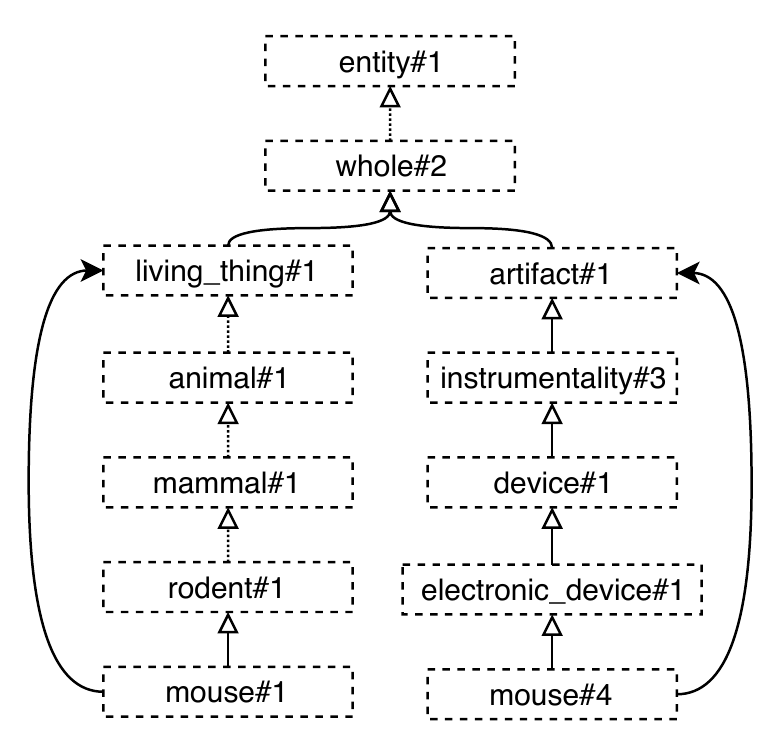}
\caption{Sense vocabulary compression trough hypernymy hierarchy applied on the first and fourth sense of the word ``mouse''. Dashed arrows mean that some nodes are skipped for clarity.}
\label{fig:mouse_hypernym}
\end{figure}

Indeed, considering a small subset of WordNet that only consists of the word ``mouse'', its first sense (the small rodent), its fourth sense (the electronic device), and all of their hypernyms. This is illustrated in \autoref{fig:mouse_hypernym}. We can see that every concept that is more specialized than the concepts ``artifact'' and ``living\_thing'' could be removed. We could map every tag of ``mouse\#1'' to the tag of ``living\_thing\#1'' and we could still be able to disambiguate this word, but with a benefit: all other ``living things'' and animals in the sense annotated data could be tagged with the same sense. They would give examples of what is an animal and then show how to differentiate the small rodent from the hand-operated electronic device. 

Therefore, the goal of our method is to map every sense of WordNet to its highest ancestor in the hypernymy hierarchy, but with the following constraints:
First, this ancestor must discriminate all the different senses of the target word.
Second, we need to preserve the hypernyms that are indispensable to discriminate the senses of the other words in the dictionary. For instance, we cannot map ``mouse\#1'' to ``living\_thing\#1", because the more specific tag ``animal\#1'' is essential to distinguish the two senses of the word ``prey'' (one sense describes a person, the other describes an animal).

\noindent Our method thus works in two steps:
\begin{enumerate}[leftmargin=*,topsep=0pt,itemsep=0pt,parsep=0pt,partopsep=0pt]
    \item We mark as ``necessary'' the children of the first common ancestor of every pair of senses of every word of WordNet.
    \item We map every sense to its first ancestor in the hypernymy hierarchy that has been previously marked as ``necessary''.
\end{enumerate}

\noindent As a result, the most specific synsets of the tree that are not indispensable for discriminating any word of the lexical inventory are automatically removed from the vocabulary. In other words, the set of synsets that is left in the vocabulary is the smallest subset of all synsets that are necessary to distinguish every sense of every word of WordNet, following the hypernym and hyponym links.

\subsection{Compression through all semantic relationships}

In addition to hypernymy and hyponymy, WordNet contains several other relationships between synsets, such as the instance relationship (e.g. ``Albert Einstein'' is an instance of ``physicist'), the meronymy (X is part of Y, or X is a member of Y) and its counterpart the holonymy, the antonymy (X is the opposite of Y), etc.

We hence propose a second method for sense vocabulary compression, that considers all the semantic relationships offered by WordNet, in order to form clusters of related synsets.

For instance, using all semantic relationships, we could form a cluster containing ``physicist'', ``physics'' (domain category), ``Albert Einstein'' (instance of), ``astronomer'' (hyponym), but also further related senses such as ``photon'', because it is a meronym of ``radiation'', which is a hyponym of ``energy'', which belongs to the same domain category of ``physics''.  

Our method works by constructing these clusters iteratively: First, we initialize the set of clusters $C$ with one synset in each cluster.
\begin{align}
    C=&\{c_0, c_1, ..., c_n\} \quad \quad S=\{s_0, s_1, ..., s_n\} \nonumber \\
    C=&\{\{s_0\},\{s_1\},...,\{s_n\}\} \nonumber
\end{align}
Then at each step, we sort $C$ by sizes of clusters, and we peek the smallest one $c_x$ and the smallest related cluster to $c_x$, $c_y$. We define a cluster being related to another if they contain at least one synset that have a semantic link together.
We merge $c_x$ and $c_y$ together, and we verify that the operation still allows to discriminate the different senses of all words in the lexical database. If it is not the case, we cancel the merge and we try another semantic link. If no link is possible, we try to create one with the next smallest cluster, and if no further link can be created, the algorithm stops.

\begin{figure}[htbp]
\centering
\includegraphics[width=1.0\linewidth]{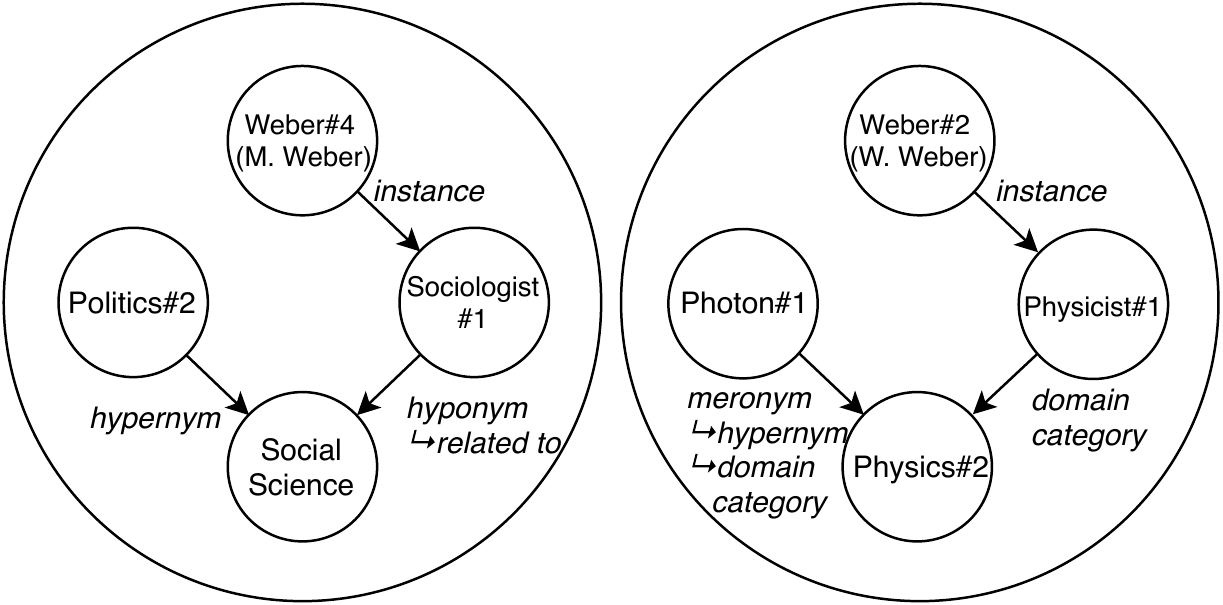}
\caption{Example of clusters of sense that could result from our method, if we limit our view to two senses of the word ``Weber'' and only some relationship links.}
\label{fig:weber_cluster}
\end{figure}

In \autoref{fig:weber_cluster}, we show a possible set of clusters that could result from our method, focusing on two senses of the word ``Weber'' and only on a few relationships.

This method produces clusters significantly larger than the method based on hypernyms. On average, a cluster has 5 senses with the hypernym method, whereas it has 17 senses with this method. This method, unlike the previous one, is also stochastic, because the formation of clusters depends on the underlying order of iteration when multiple clusters are the same size. However, because we always sort clusters by size before creating a link, we observed that the final vocabulary size (i.e. number of clusters) is always between 11\,000 and 13\,000.
In the following, we consider a resulting mapping where the algorithm stopped after 105\,774 steps.

\begin{table}[htbp]
\small
\begin{center}
\tabulinesep=3pt
\taburowcolors[2]2{white..gray!10}
\begin{tabu} to \linewidth {|X[2.9lm]|X[1.9cm]|X[1.8cm]|X[1.8cm]|} \firsthline
Method & Vocabulary size & Compres\-sion~rate & SemCor Coverage  \\
\hline
No~compression & 206\,941 & 0\% & 16\% \\
Synonyms & 117\,659 & 43\% & 22\% \\
Hypernyms & 39\,147 & 81\% & 32\% \\
All relations & 11\,885 & 94\% & 39\% \\
\lasthline
\end{tabu}
\end{center}
\caption{Effects of the sense vocabulary compression on the vocabulary size and on the coverage of the SemCor.}
\label{tab:vocabulary_compression}
\end{table}

In \autoref{tab:vocabulary_compression}, we show the effect of the common compression through synonyms,
our first proposed compression through hypernyms, and our second method of compression through all semantic relationships, on the size of the vocabulary of WordNet sense tags, and on the coverage of the SemCor corpus. 
As we can see, the sense vocabulary size is drastically decreased, and the coverage of the same corpus really improved.

\section{Experiments}

In order to evaluate our sense vocabulary compression methods, we applied them on a neural WSD system 
based on a softmax classifier capable of classifying a word in all possible synsets of WordNet (see \autoref{sec:classif_based_wsd}).

We implemented a system 
similar to \citet{raganato2017}'s BiLSTM but with some key differences. In particular, 
we used BERT contextualized word vectors \citep{devlin2018bert} in input of our network,
Transformer encoder 
layers 
\citep{vaswani2017} instead of LSTM 
layers 
as hidden units, 
our output vocabulary only consists of sense tags seen during training (mapped according to the compression method used), 
and we ignore the network's predictions on words that are not annotated.

\subsection{Implementation details}

For BERT, we used the model named ``bert-large-cased'' of the PyTorch implementation\footnote{\url{https://github.com/huggingface/pytorch-pretrained-BERT}}, which consists of vectors of dimension 1024, trained on BooksCorpus and English Wikipedia.

Due to the fact that BERT's internal tokenizer sometimes split words in multiples tokens (i.e. \lbrack{}``rodent''\rbrack{} becomes \lbrack{}``rode'', ``\#\#nt''\rbrack{}), we trained our system to predict a sense tag on the first token only of a splitted annotated word.

For the Transformer encoder layers, we used the same parameters as the ``base'' model of \citet{vaswani2017}, that is 6 layers with 8 attention heads, a hidden size of 2048, and a dropout of 0.1.

Finally, because BERT already encodes the position of the words inside their vectors, we did not add any positional encoding.

\subsection{Training}

We compared our sense vocabulary compression methods on two training sets: 
The SemCor, and  
the concatenation of the SemCor and the Princeton WordNet Gloss Corpus (WNGC).
The latter is a corpus distributed as part of WordNet since its version 3.0, and it consists of the definitions (glosses) of every synset of WordNet, with 
words manually or semi-automatically sense annotated.
We used the version of these corpora given as part of the UFSAC~2.1 resource\footnote{\url{https://github.com/getalp/UFSAC}} 
\citep{vialhal01718237}.

We performed every training for 20 epochs. 
At the beginning of each epoch, we shuffled the training set. We evaluated our model at the end of every epoch on a development set,
and we kept only the one which obtained the best F1 WSD score. 
The development set was composed of 
4\,000 random sentences taken from the Princeton WordNet Gloss Corpus for the models trained on the SemCor, and 4\,000 random sentences extracted from the whole training set for the other models.

\noindent For each training set, we trained three systems: 
\begin{enumerate}[leftmargin=*,topsep=0pt,itemsep=0pt,parsep=0pt,partopsep=0pt]
    \item 
A ``baseline'' system that predicts a tag belonging to all the synset tags seen during training,
thus using the common 
vocabulary compression through synonyms method.
    \item 
A ``hypernyms'' system which applies our vocabulary compression through hypernyms algorithm on the training corpus.
    \item
A ``all relations'' system which applies our second vocabulary compression through all relations on the training corpus.
\end{enumerate}

\noindent We trained with mini-batches of 100 sentences, truncated to 80 words, 
and we used Adam \cite{KingmaB14} with a learning rate of $0.0001$ as the optimization method.

\begin{table}[htbp]
\small
\tabulinesep=3pt
\begin{tabu} to \linewidth {|X[1lm]|X[0.8cm]|X[1.2cm]|}
\firsthline
System & SemCor & SemCor+WNGC \\
\hline
baseline & 77.15M & 120.85M  \\
\rowcolor{gray!10} 
hypernyms & 63.44M & 79.85M \\
all relations & 55.16M & 60.27M  \\
\lasthline
\tabuphantomline
\end{tabu}
\caption{Number of parameters of neural models.}
\label{tab:times}
\end{table}

All models have been trained on one Nvidia's Titan X GPU. 
The 
number of parameters of individual models
are displayed in 
\autoref{tab:times}. As we can see, our compression methods
drastically reduce the number of parameters, by a factor of 1.2 to 2.

\subsection{Evaluation}

\begin{table*}[htbp]
\small
\begin{center}
\tabulinesep=3pt
\taburowcolors[3]2{white..gray!10}
\begin{tabu} to \linewidth {|X[26lm]|X[4cm]X[4cm]X[4cm]X[4cm]X[4cm]|X[4cm]X[4cm]X[4cm]X[4cm]|X[4cm]|X[4cm]|} \firsthline
 & SE2 & SE3 & SE07 & SE13 & SE15 & \multicolumn{5}{c|}{ALL (concat. of previous tasks)} & SE07\\
System &  &  & 17 &  &  & nouns & verbs & adj. & adv. & total & 07\\
\hline
First sense baseline & 65.6 & 66.0 & 54.5 & 63.8 & 67.1 & 67.7 & 49.8 & 73.1 & 80.5 & 65.5 & 78.9\\
\hline
HCAN \citep{luo2018b} & 72.8 & 70.3 & - & 68.5 & 72.8 & 72.7 & 58.2 & 77.4 & 84.1 & 71.1 & - \\
LSTMLP \citep{yuan_2016} & 73.8 & 71.8 & 63.5 & 69.5 & 72.6 & $\dagger$73.9 & - & - & - & $\dagger$71.5 & 83.6 \\
\hline
SemCor, baseline & 77.2 & 76.5 & 70.1 & 74.7 & 77.4 & 78.7 & 65.2 & 79.1 & 85.5 & 76.0 & 87.7 \\
SemCor, hypernyms & 77.5 & 77.4 & 69.5 & 76.0 & 78.3 & 79.6 & 65.9 & 79.5 & 85.5 & 76.7 & 87.6 \\
SemCor, all relations & 76.6 & 76.9 & 69.0 & 73.8 & 75.4 & 77.2 & 66.0 & 80.1 & 85.0 & 75.4 & 86.7 \\
SemCor+WNGC, baseline & \tbf{79.7} & 76.1 & \tbf{74.1} & 78.6 & 80.4 & 80.6 & 68.1 & 82.4 & 86.1 & 78.3 & 90.4 \\
SemCor+WNGC, hypernyms & \tbf{79.7} & 77.8 & 73.4 & \tbf{78.7} & \tbf{82.6} & 81.4 & 68.7 & 83.7 & 85.5 & \tbf{79.0} & 90.4 \\
SemCor+WNGC, all relations & 79.4 & \tbf{78.1} & 71.4 & 77.8 & 81.4 & 80.7 & 68.6 & 82.8 & 85.5 & 78.5 & \tbf{90.6} \\
\lasthline
\rowcolor{white}\tabuphantomline
\end{tabu}
\end{center}
\vspace{-4pt}
\caption{F1 scores (\%) on the English WSD tasks of the evaluation campaigns 
SensEval/SemEval. The task ``ALL'' is the concatenation of SE2, SE3, SE07~17, SE13 and SE15.
The 
first sense 
is 
assigned 
on words for which none of its sense has been observed during the training. 
Results in \tbf{bold} are to our knowledge the best results obtained on the task.
Scores prefixed by a dagger ($\dagger$) are not provided by the authors but are deduced from their other scores.
}
\label{tab:scores}
\end{table*}

We evaluated our models on 
all evaluation corpora commonly used in WSD, that is the English all-words WSD tasks 
of the evaluation campaigns SensEval/SemEval. We used the fine-grained evaluation corpora from the evaluation framework of \citet{raganatocamachocolladosnavigli2017},
which consists of SensEval~2 \citep{Edmonds2001}, SensEval~3 \citep{W040811}, SemEval~2007 task~17 \citep{Pradhan2007}, SemEval~2013 task~12 \citep{Navigli2013} and SemEval~2015 task~13 \citep{moronavigli2015}, as well as 
the ``ALL'' corpus consisting of the concatenation of all previous ones. 
We also compared our result on the coarse-grained task~7 of SemEval~2007 \citep{Navigli2007} which is not present in this framework. 

For each evaluation, we trained 8 independent models, and we give 
the score obtained by an ensemble system that
averages their predictions
through a geometric mean.

\begin{table}[htbp]
\small
\centering
\tabulinesep=3pt
\setlength\tabcolsep{1.5pt}
\taburowcolors[2]2{white..gray!10}
\begin{tabu} to \linewidth {|X[1.6lm]|X[0.65cm]|X[0.65cm]|}
\firsthline
System & No Backoff & Backoff on Monosemics \\
\hline
SemCor, baseline & 93.23\% & 98.13\% \\
SemCor, hypernyms & 98.75\% & 99.68\% \\
SemCor, all relations & 99.67\% & 99.99\% \\
SemCor+WNGC, baseline & 98.26\% & 99.41\% \\
SemCor+WNGC, hypernyms & 99.83\% & 99.96\% \\
SemCor+WNGC, all relations & 99.99\% & 100\% \\
\lasthline
\tabuphantomline
\end{tabu}
\caption{Coverage of our systems on the task ``ALL''. ``Backoff on Monosemics'' means that monosemic words are considered annotated.}
\label{tab:coverage}
\end{table}

In the results in \autoref{tab:scores}, we first observe that 
our systems that use the sense vocabulary compression through hypernyms or through all relations obtain scores 
that are overall equivalent to the systems that do not use it.

Our methods greatly improves their coverage on the evaluation tasks however.
As we can see in \autoref{tab:coverage},
on the total of 7\,253 words to annotate for the corpus ``ALL'', the baseline system trained on the SemCor is not able to annotate 491 of them, while the vocabulary compression through hypernyms reduces this number to 91 and 24 for the compression through all relations. 

\begin{table*}[htbp]
\small
\begin{center}
\tabulinesep=2pt
\taburowcolors[3]2{white..gray!10}
\begin{tabu} to \linewidth {|X[5lm]|X[3cm]|X[2cm]|X[1cm]|X[1cm]||X[1cm]|X[1cm]||X[1cm]|X[1cm]|}
\firsthline
\multirow{3}{*}{Training Corpus} & \multirow{3}{*}{Input Embeddings} & \multirow{3}{*}{Ensemble} & \multicolumn{6}{c|}{F1 Score on task ``ALL'' (\%)} \\ 
 &  &  & \multicolumn{2}{c||}{Baseline} & \multicolumn{2}{c||}{Hypernyms} & \multicolumn{2}{c|}{All relations} \\
 & & & $\Bar{x}$ & $\sigma$ & $\Bar{x}$ & $\sigma$ & $\Bar{x}$ & $\sigma$ \\
\hline
SemCor+WNGC & BERT & Yes & 78.27 & - & 79.00 & - & 78.48 & - \\
SemCor+WNGC & BERT & No & 76.97 & $\pm 0.38$ & 77.08 & $\pm 0.17$ & 76.52 & $\pm 0.36$ \\
SemCor+WNGC & ELMo & Yes & 75.16 & - & 74.65 & - & 70.58 & - \\
SemCor+WNGC & ELMo & No & 74.56 & $\pm 0.27$ & 74.36 & $\pm 0.27$ & 68.77 & $\pm 0.30$ \\
SemCor+WNGC & GloVe & Yes & 72.23 & - & 72.74 & - & 71.42 & - \\
SemCor+WNGC & GloVe & No & 71.93 & $\pm 0.35$ & 71.79 & $\pm 0.29$ & 69.60 & $\pm 0.32$ \\
SemCor & BERT & Yes & 76.02 & - & 76.73 & - & 75.40 & - \\
SemCor & BERT & No & 75.06 & $\pm 0.26$ & 75.59 & $\pm 0.16$ & 73.91 & $\pm 0.33$ \\
SemCor & ELMo & Yes & 72.55 & - & 73.09 & - & 69.43 & - \\
SemCor & ELMo & No & 72.21 & $\pm 0.13$ & 72.83 & $\pm 0.24$ & 68.74 & $\pm 0.29$ \\
SemCor & GloVe & Yes & 70.77 & - & 71.18 & - & 68.44 & - \\
SemCor & GloVe & No & 70.51 & $\pm 0.16$ & 70.77 & $\pm 0.21$ & 67.48 & $\pm 0.55$ \\
\hline
\multicolumn{9}{|l|}{\quad \quad \quad  HCAN \cite{luo2018b} (fully reproducible state of the art)} \\
SemCor+WordNet glosses & GloVe & No & 71.1 & \multicolumn{5}{c|}{}  \\
\hline
\multicolumn{9}{|l|}{\quad \quad \quad LSTMLP \cite{yuan_2016} (state of the art scores but use private data)} \\
SemCor+1K (private) & private & No & 71.5 & \multicolumn{5}{c|}{} \\
\lasthline
\rowcolor{white}\tabuphantomline
\end{tabu}
\end{center}
\caption{Ablation study on the task ``ALL'' (i.e. the concatenation of all SensEval/SemEval tasks). For systems that do not use ensemble, we display the mean score ($\Bar{x}$) of eight individually trained models along with its standard deviation ($\sigma$).}
\label{tab:ablation}
\end{table*}

When adding the Princeton WordNet Gloss Corpus to the training set, only one word (the monosemic adjective ``cytotoxic'') cannot be annotated with the system that uses the compression through all relations because its sense has not been observed during training.

If we exclude the monosemic words, the system based on our compression method through all relations miss only one word (the adverb ``eloquently'') when trained on the SemCor, and has a coverage to 100\% when the WNGC is addded.

In comparison to the other works, 
thanks to the Princeton WordNet Gloss Corpus added to the training data and the use of BERT as input embeddings, 
we outperform systematically the state of the art on every task.
\subsection{Ablation Study}

In order to give a better understanding of the origin of our scores, we provide a study of the impact of our main parameters on the results. In addition to the training corpus and the vocabulary compression method, we chose two parameters that differentiate us from the state of the art:
the pre-trained word embeddings model
and the ensembling method,
and we have made them vary. 

For the word embeddings model, we experimented with BERT \citep{devlin2018bert} as in our main results, with ELMo \citep{Peters2018}, and with GloVe \citep{pennington2014glove}, the same pre-trained word embeddings used by \citet{luo2018b}. 
For ELMo, we used the model trained on Wikipedia and the monolingual news crawl data from WMT 2008-2012.\footnote{\url{https://allennlp.org/elmo}} 
For GloVe, we used the model trained on Wikipedia 2014 and Gigaword 5.\footnote{\url{https://nlp.stanford.edu/projects/glove/}}
Due to the fact that GloVe embeddings do not encode the position of the words (a word has the same vector representation in any context), we used bidirectional LSTM cells of size 1\,000 for each direction, instead of Transformer encoders for this set of experiments. In addition, because the vocabulary of GloVe is finite and all words are lowercased, we lowercased the inputs, and we assigned a vector filled with zeros to out-of-vocabulary words.

For the ensembling method, we either perform ensembling as in our main results, by averaging the prediction of 8 models trained separately or 
we give the mean and the standard deviation of the scores of the 8 models evaluated separately.   

As we can see in \autoref{tab:ablation}, the additional training corpus (WNGC) and even more the use of BERT as input embeddings both have a major impact on our results and lead to scores above the state of the art. 
Using BERT instead of ELMo or GloVe improves respectively the score by 
approximately 3 and 5 points in every experiment, and adding the WNGC to the training data improves it by approximately 2 points. Finally, using ensembles adds roughly another 1 point to the final F1 score.

Finally, through the scores obtained by invidual models (without ensemble), we can observe on the standard deviations that the vocabulary compression method through hypernyms never impact significantly the final score. However, the compression method through all relations seems to negatively impact the results in some cases (when using ELMo or GloVe especially).

\section{Conclusion}
{
In this paper, we presented two new methods that improve the coverage and the capacity of generalization of supervised WSD systems, by narrowing down the number of different sense in WordNet in order to keep only the senses that are essential for differentiating the meaning of all words 
of the lexical database.
On the scale of the whole lexical database, we showed that these methods can shrink the total number of different sense tags in WordNet to only 6\% of the original size, and that the coverage of an identical training corpus has more than doubled.
We implemented a state of the art WSD neural network and we showed that these methods 
compress the size 
of the underlying models by a factor of 1.2 to 2, and greatly improve their coverage on the evaluation tasks. As a result, we reach a coverage of 99.99\% of the evaluation tasks (1 word missing on 7\,253) when training a system on the SemCor only, and 100\% when adding the WNGC to the training data, on the polysemic words. Therefore, the need for a backoff strategy is nearly eliminated.
Finally, our method combined with the recent advances in contextualized word embeddings and with a training corpus composed of sense annotated glosses, our system achieves scores that 
considerably outperform the state of the art on all WSD evaluation tasks.

}

\bibliography{biblio}
\bibliographystyle{acl2010}

\end{document}